\documentclass{article}
\usepackage{epsfig}
\usepackage[cp1250]{inputenc}
\usepackage{dsfont}
\usepackage{amsfonts}
\usepackage{amsmath}
\usepackage{subfigure}

\newcommand {\R} {\mathds{R}}

\newcommand {\E} {\mathds{E}}
\newcommand {\e} {\varepsilon}

\newcommand {\rr} [2] {\frac{\partial #1}{\partial #2}}
\newcommand {\rrr} [3] {\frac{\partial^2 #1}{\partial #2\partial #3}}
\newcommand {\pr} [1] {\frac{\partial}{\partial #1}}

\newcommand {\Id}{\mathop{\mathrm{Id}}}
\newcommand {\EI}{\mathop{\mathrm{EI}}}
\newcommand {\REI}{\mathop{\mathrm{REI}}}

\newtheorem {example}{Example}

\newcommand {\Smieszne}[1]{#1}
\newcommand {\Wyciete}[1]{}

\newcommand {\aposteriori}{{\it a posteriori}}

\title{Relative Expected Improvement in Kriging Based Optimization}

\author{Łukasz Łaniewski-Wołłk \\ \it Institute of Aeronautics and Applied Mechanics\\ Warsaw University of Technology \\
Nowowiejska 24, 00-665 Warsaw, Poland \\
e-mail: llaniewski@meil.pw.edu.pl \\
Web page: http://c-cfd.meil.pw.edu.pl/}

\begin{document}
\maketitle

\abstract{We propose an extension of the concept of Expected
Improvement criterion commonly used in Kriging based optimization.
We extend it for more complex Kriging models, e.g. models using
derivatives. The target field of application are CFD problems, where
objective function are extremely expensive to evaluate, but the
theory can be also used in other fields.}

\section{INTRODUCTION}

Global optimization is a common task in advanced engineering. The
objective function can be very expensive to calculate or measure. In
particular this is the case in Computational Fluid Dynamics (CFD)
where simulations are extremely expensive and time-consuming. At
present, the CFD code can also generate the exact derivatives of the
objective function so we can use them in our models. The long
computation to evaluate the objective function and (as a rule) high
dimension of the design space make the optimization process very
time-consuming.

Widely adopted strategy for such objective functions is to use
response function methodology. It is based on constructing an
approximation of the objective function based on some measurements
and subsequently finding points of new measurements that enhance our
knowledge about the location of optimum.

One of the commonly used response functions models is the Kriging
model \cite{Sacks1989,Leary2004,Morris1993,Jones1998}. This
statistical estimation model considers the objective function to be
a realization of a random field. We can construct a least square
estimator. If we assume the field to be gaussian, the least square
estimator is the Bayesian estimator. Conditional distribution of the
field with respect to the measurements (\aposteriori) is also
gaussian with known both mean and covariance.

One of the methods to find a point for new measurement is the
Expected Improvement criterion\cite{Jones1998}. It uses a Expected
Improvement function: \[\EI(x) = \E( \min{(\hat{F}_{\min}, F(x))}
)\] where $F$ is the \aposteriori\ field and $\hat{F}_{\min}$ is the
minimum of estimator. The new point of measurement is chosen in the
minimum of $\EI$ function.

Many modifications and enhancements were considered for the Kriging
model. Application of linear operators, e.g. derivatives, integrals
and convolutions, are easy to incorporate in the
model\cite{Leary2004,Morris1993}. \Wyciete{We can use derivatives,
because solution of the adjoint (dual) equation provides a method of
calculating all the derivatives at once, with the cost comparable to
calculating of the objective. We can now add the information on the
gradient to the Kriging model. Furthermore we can enhance our model
with high and low fidelity models as well as include convolution to
obtain a robust response.}

Each of these extensions of classic Kriging model is based on
measuring something else then is returned as the response. For
example we measure gradient and value of the function, but the
response is only the function. The Expected Improvement states that
we should measure the function in place where the minimum of
response can be mostly improved. But for classic model the notion of
the measured and the response functions are the same.

The purpose of this paper is the investigation wether the concept of
$\EI$ can be extended for enhanced Kriging models.

\section{RELATIVE EXPECTED IMPROVEMENT}
\subsection{Efficient Global Optimization}
Jones et al.\cite{Jones1998} propose an Efficient Global
Optimization (EGO) algorithm based on Kriging model and Expected
Improvement. It consists of the following steps:
\begin{enumerate}
\item Select a learning group $x_1, \ldots, x_n$. Measure objective
function $f$ in these points $f_i = f(x_i)$.
\item \label{EGO petla} Construct a Kriging approximation $\hat{F}$
based on measurements $f_1, \ldots, f_n$.
\item Find the minimum of $\EI(x)$ function for the approximation.
\item Augment $n$ and set $x_{n}$ at the minimum of $\EI$.
\item Measure $f_n=f(x_n)$ and go back to \ref{EGO petla}
\end{enumerate}
$\EI$ function can have many local minima (is highly multi-modal)
and is potentially hard to minimize. The original paper proposed
Branch and Bound Algorithm (BBA) to efficiently optimize the $\EI$
function. To use BBA authors had to establish upper and lower bonds
on minimum of $\EI$ function over a region. It was fairly easy and
was the main source of effectiveness of EGO. While proposing an
extension of $\EI$ concept we also have to propose a suitable
methods of it's optimization.

\subsection{Gaussian Kriging}
Kriging, is a statistical method of approximation a
multi-dimensional function basing on values in a set of points. The
Kriging estimator (approximation) can be interpreted as a
least-square estimator, but also as a Bayes estimator. We will use
the latter interpretation as in the original $\EI$ definition.

Let us take an objective function $f:\Omega \rightarrow \R$. For
some probabilistic space $\left(\Gamma, \mathcal{F},
\mathds{P}\right)$, we consider a random gaussian field  $F$ on
$\Omega$ with the known mean $\mu$ and covariance $K(x,y)$. Now we
take a measurements of the objective at points $x_1, \ldots, x_n$ as
$f_i = f(x_i)$. The Bayes estimator of $f$ is:
\[\hat{F}(x) = \E\left(F(x)\mid\forall_i\;F(x_i)=f_i\right)\]
Where $\E(A\mid B)$ is conditional expected value of $A$ with
respect to $B$. This estimator at $y$ will be called the response at
$y$ and the $(x_i,f_i)$ pairs will be called measurements at $x_i$.

Let us take an event
$M=\left\{\forall_i\;F(x_i)=f_i\right\}\subset\Gamma$ and a
\aposteriori\ probability space $\left(M,\mathcal{F}_M,
\mathds{P}(\cdot\mid M)\right)$. $\E_M$ will stand for expected
value in \aposteriori. Field $F$ considered on the $M$ space is also
a gaussian random field with known both mean $\mu_M$ and covariance
$K_M$. We will call this field, the \aposteriori\ field.

\subsection{From $\EI$ to $\REI$}
We would want to estimate how much the minimum of $\hat{F}$ we will
be improved if will measure $f$ at some point. Estimator $\hat{F}$
after the measurement in $x$ can be writhen as $F_x = \E_M(F|F(x))$.
The best estimate of the effect would be $\E_M \inf_\Omega F_x$. But
computing it would be very time-consuming. The idea of Expected
Improvement ($\EI$) is to take
\[\EI(x)=\E_M \min\{F_{\min}, F(x)\}\] where $F_{\min}$ is the
actual minimum of approximation $\hat{F}$. Expected Improvement is
in fact expected value of how response at $x$ will improve the
actual minimum of $\hat{F}$. Of course the definition is equivalent
to:
\[\EI(x)=\E_M \min\{F_{\min}, \E_M(F(x)|F(x))\}\] This formulation has a natural extension. Let
us define, for a set of points $\eta=\{\eta_1,\ldots,\eta_l\}$, a
augmented estimator $F_\eta(x) = \E_M \left(F(x)\mid F(\eta_1),
\ldots,F(\eta_l)\right)$. For another set of points
$\zeta=\{\zeta_1,\ldots,\zeta_k\}$ we can define:
\[\REI(\zeta,\eta)=\E_M \min\{F_{\min}, F_\eta(\zeta_1), \ldots, F_\eta(\zeta_k)\}\]
Our Relative Expected Improvement ($\REI$) is the expected value of
how much the response at $\zeta$ will improve the minimum of
$\hat{F}$ if we measure at $\eta$. This definition implies
$\REI(\{x\},\{x\}) = \EI(x)$.

We can use also a more general version:
\[{\REI}_m(\zeta,\eta)=\E_M \min\{F_\eta(\zeta_1), \ldots, F_\eta(\zeta_k)\}\]

The main advantage of $\REI$ function is that we can examine the
response in a different region then the region of acceptable
measurements. A simple example illustrates it very well:

\Smieszne{\begin{figure}[h]
\begin{center}
\epsfig{file=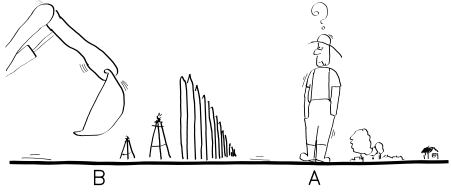}\caption{$A$ and $B$ sets of possible drilling
points}
\end{center}
\end{figure}
}

\begin{example}
We're searching for some mineral. We have to estimate the maximum
mineral content in somebody's land before buying it. We cannot drill
at his estate, but we can drill everywhere around it.
\end{example}
In this example response and measurements are in a different
regions, so we cannot use $\EI$. If the estate is $A$ and the
surrounding ground is $B$, in order to find the best place to drill,
we would have to search for the minimum of $\REI(\{x\},\{y\})$ for
$x\in A$ and $y\in B$.

\section{APPLICATION}
\subsection{Populations of measurement points}
The first application of using $\REI$ instead of $\EI$ is when we
want to find a collection of measurement points instead of a single
point, e.g. when the objective function can be computed
simultaneously at these points. It's a possibility of making the
optimization process more parallel.

\Smieszne{\begin{figure}[h]
\begin{center}
\subfigure[]{\epsfig{file=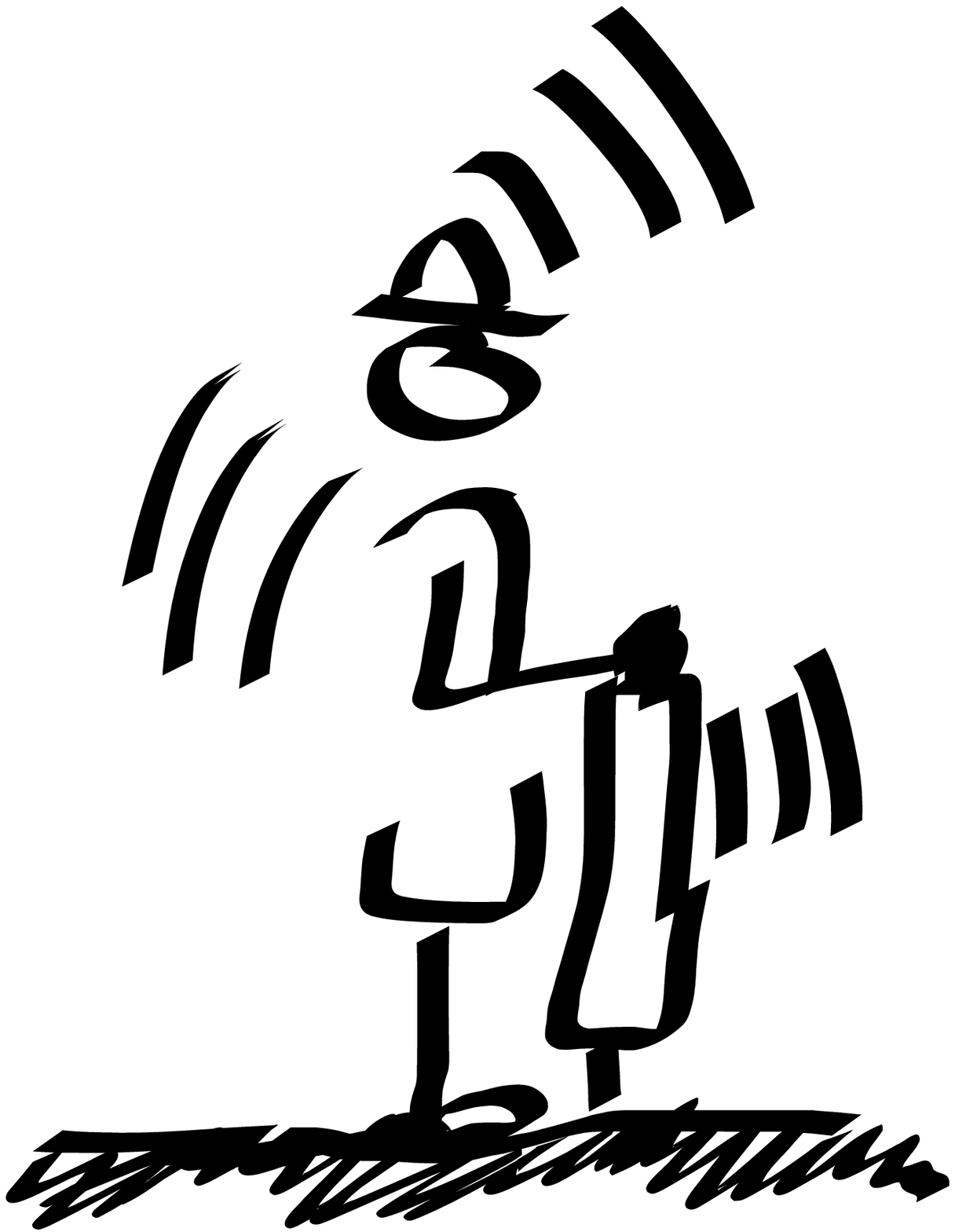, height=1in}\label{one
measurement fig}} \subfigure[]{\epsfig{file=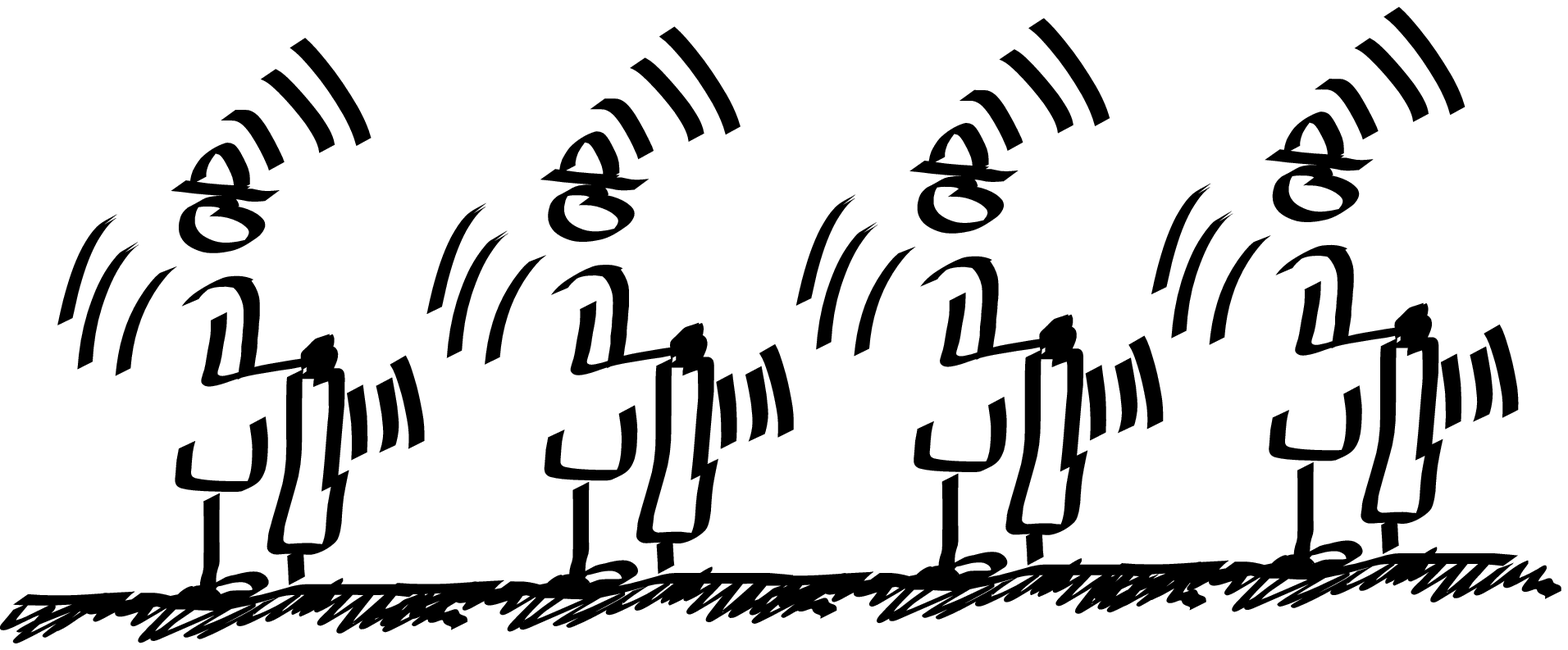,
height=1in}\label{multi measurement fig}} \caption{\subref{one
measurement fig} One point of measurement. \subref{multi measurement
fig} Population of measurement points.}
\end{center}
\end{figure}}

\begin{example}
We have $k$ processors to solve our CFD problem, each running a
separate flow case.
\end{example} This procedure could be, for example, to optimize
$\REI(\{\zeta_1, \ldots, \zeta_n\},\{\zeta_1, \ldots, \zeta_n\})$.
The main advantage in using such an expression, over using some
selection of $\EI$ minima, is that $\REI$ considers the correlation
between these points. For example, if $x$ and $y$ are strongly
correlated, we don't want to measure in both these points, because
the value in $x$ implies the value in $y$.

\subsection{Input enhancements}
The other application field is enhancing the Kriging model, by some
other accessible information than the values in points.

Let us define a generalized point as a pair $(x,P)$, where
$x\in\Omega$ is a point, and $P$ is a linear operator. We can say
that $f(x,P) = (P f)(x)$. The field $F(x,P)$ is also gaussian with:
\[\mu(x,P) = (P\mu)(x)\]
\[K(x,P;y,S) = P_xS_yK(x,y)\]
where $P_x$ stands for applying $P$ to $K$ as a function of the
first coefficient. Now all the earlier definitions can be extended
to generalized points. (In fact this enhancement can be done by
enlarging $\Omega$ to $\Omega\times \{\Id, P, S, \ldots\}$)

\begin{example}
The CFD code is solving the main and the adjoint problem. We have
both the value of our objective and its derivatives with respect to
design parameters. We want to find the best place to measure these
values.
\end{example}

We can use $f(x,\pr{x_k}) = \rr{f}{x_k}(x)$ to interpret measuring
the derivatives of $f$ interpret as measuring at points
$(x,\pr{x_k})$. In the example we have not only calculated the value
at $(x,\Id)$, but also at $(x,\pr{x_k})$. If we have $d$ design
parameters (that is $\Omega\subset\R^d$) we have $d+1$ measurements
simultaneously. We can now optimize:
\[\REI\left(\left\{(\zeta_1,\Id),\ldots,(\zeta_{d+1},\Id)\middle\},\middle\{(x,\Id),(x,\pr{x_1}),\ldots,(x,\pr{x_d})\right\}\right)\]
We take $d+1$ points of response $\zeta$ to maximize the effect of
all the measured derivatives. We could of course use $\EI$. In that
case we would select the next point as if we're measuring only the
value. By using $\REI$ we're incorporating the derivative
information not only in the model, but also in the selection
process. The disadvantage of such an expression is that we search in
$\Omega^{d+2}$ which is $d\cdot(d+2)$-dimensional.

\subsection{Multi-effect response}

Next on our list is the multi-effect model. We can imagine that our
measured function is composed of several independent or dependent
effects, while our objective function is only one of them. The
simplest case is when we want to optimize objective which we
measuring with an unknown error.

Let us now say that $F$ consists of several components $F(x) =
(Z(x), W(x), V(x), \ldots)$. Same letters will stand for linear
operators, such that $F(x,Z)=Z(x)$.

\begin{example} Suppose that we're searching for mineral $A$, but our drilling equipment
for measuring content of $A$, cannot distinguish it from another
mineral $B$. We know on the other hand that the latter is
distributed randomly and in small patches.
\end{example}
Let $Z$ be our objective function (mineral $A$ content) and $\e$ a
spatially-correlated error (mineral $B$ content). We can measure
only $Z+\e$ while we want to optimize $Z$. In this example we can
optimize $\REI(\{(x,Z)\},\{(x,Z+\e)\})$. Such a procedure will
simultaneously take into consideration optimization of the objective
and correction of the error. To fully understand why this example is
important, we have to remember that drilling in the same place twice
would give the same result. The error correction in our procedure
will bear this in mind and will avoid duplication of measurements.

This model would include results obtained from lower-quality
numerical calculations. For an iterative algorithm (non-random), we
can state a higher error bound and reduce the number of iterations.
We cannot assume the error to be fully random, because starting from
the same parameters, the algorithm will give the same results.
That's why a good Kriging model, would recognize the error to be a
narrowly correlated random field $\e$.

\Smieszne{
\begin{figure}[h]
\begin{center}
\subfigure[]{\epsfig{file=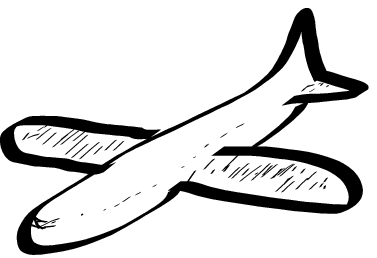}\label{high fidel fig}}
\subfigure[]{\epsfig{file=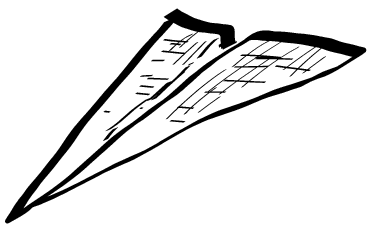}\label{low fidel fig}}
\caption{\subref{high fidel fig} High and \subref{low fidel fig} low
fidelity models}
\end{center}
\end{figure}
}

\begin{example} We have two CFD models. One accurate and the other approximate,
but very fast (high and low-fidelity models). We know also, that the
low-fidelity model is ``smoothe'' with respect to the design
parameters.
\end{example}
Let $Z$ be our objective function and $W$ be a approximation of $Z$.
In this example we can separately optimize:
\begin{eqnarray*}\REI(\{(\zeta_1,Z),\ldots,(\zeta_k,Z)\},\{(x,Z)\})\\
\REI(\{(\zeta_1,Z),\ldots,(\zeta_k,Z)\},\{(x,W)\})\end{eqnarray*}
and subsequently choose between these two points. Field $W$ is
strongly spatially-correlated (``smooth'') and as such it's
measurement can have wider effect than $Z$.  We can also take in to
consideration the cost of the computation and select a better {\it
improvement-to-cost} ratio.

\subsection{Robust response}

The last field of application, that we will discus, is the robust
response. If for instance after optimization, the optimal solution
will be used to manufacture some objects, we can be sure that the
object will be manufactured within certain tolerance. In other
words, if the selected point is $x$, the actual point will be
$x+\epsilon$. Our real objective function is the average performance
of these $x+\epsilon$.

\Smieszne{
\begin{figure}[h]
\begin{center}
\subfigure[]{\epsfig{file=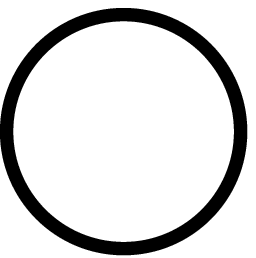}\label{designed product
fig}} \subfigure[]{\epsfig{file=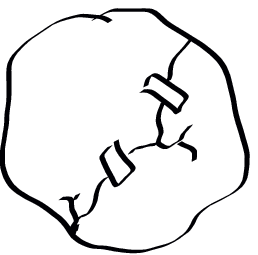}\label{manufactured
product fig}} \caption{\subref{designed product fig} Designed and
\subref{manufactured product fig} manufactured product}
\end{center}
\end{figure}
}

\begin{example}Suppose we can calculate the drag force of a car. Our factory, makes cars with
some known accuracy. We want to find the car shape, that will give
the lowest average drag when made in our factory.\end{example}

Let $Z$ be our objective function and $\epsilon$ - the manufacturing
error. We can measure only $Z(x)$ while we want to optimize $\E
Z(x+\epsilon)$. Let us say that $\epsilon$ is a random variable (for
instance $N(0,\Sigma)$), and let $\phi_\epsilon$ be it's probability
density. Now $\E(h(x+\epsilon)) = (\phi_\epsilon \ast h) (x) =
h(x,\phi_\epsilon \ast)$. In above example we can use:
\[\REI\left(\left\{(\zeta,\phi_\epsilon \ast Z)\middle\},\middle\{(\eta,Z)\right\}\right)\]
The robust response stated as above, has a good physical
interpretation. It is also fairly easy to use as long as we can
effectively calculate convolution of $\phi_\epsilon$ and the
covariance function.

It's also good to look at this kind of robust response, as a penalty
for the second derivative. If $\epsilon\sim N(0,\Sigma)$, then:
\[\E(h(x+\epsilon)) \simeq h(x) + \frac{1}{2}\sum_{ij}
\rrr{h}{x_i}{x_j}\Sigma_{ij}\] Of course such a penalty would also
be a linear operator $P_\Sigma h = h + \frac{1}{2}\sum_{ij}
\rrr{h}{x_i}{x_j}\Sigma_{ij}$ and as such can be used instead of
$\phi_\epsilon\ast$. This approach can be useful for convolutions
that are expensive to calculate.

\section{OPTIMIZATION}

\subsection{Upper bounds}
As Jones et al.\cite{Jones1998} noted, $\EI$ function can be highly
multi-modal and potentially hard to optimize. To use the branch and
bound algorithm (BBA), we have to establish a good upper bounds on
$\REI$.

We defined $\REI$ to be:
\[\REI(\zeta,\eta)=\E_M \min\{F_{\min}, F_\eta(\zeta_1), \ldots, F_\eta(\zeta_k)\}\]
where $F_\eta(x) = \E_M \left(F(x)\mid F(\eta_1),
\ldots,F(\eta_l)\right)$. It is clear that $F_\eta$ is a gaussian
field (in fact with only $l$ degrees of freedom). We can calculate
its mean and covariance depending on $\eta$. In such a case we would
wan't to establish upper bounds for an expresion:
\[\Psi_{\mu,\Sigma} = \E \min\{\gamma_1,\ldots,\gamma_p\}\]
for some $\gamma \sim N(\mu,\Sigma)$. To bound such an expression,
we can use recent extensions of comparison principle by
Vitale\cite{Vitale2000}. The comparison principle states that the
$\Phi_{\mu,\Sigma}$ is greater, the greater are $\E
(\gamma_i-\gamma_j)^2 = \Sigma_{ii} + \Sigma_{jj} - 2\Sigma_{ij}$.
To calculate the upper bound for $\REI$, we can maximize these
expressions over a region and then calculate the independent but
differently distributed (IDD) gaussian variables dominating $\REI$.
Construction of such dominating IDD variables is discussed in
Ross\cite{Ross2008}.

\subsection{Exact calculation}
In the last iterations of BBA the IDD-based bounds will be
insufficient. The main direction of further research will be to
establish a good method of calculating an exact bound on
$\Psi_{\mu,\Sigma}$. Actual algorithms in this field are based on
Monte Carlo or quasi-Monte Carlo methods, for instance using results
by Genz\cite{Genz1992}.

\section{CONCLUSIONS}
Relative Expected Improvement is proposed to extend the concept of
$EI$ for more complex Kriging models. It can help search for new
points of measurements and for populations of such points. It can
also help to use derivative information more efficiently. Further
research is needed to find efficient implementation of this concept.

\section{ACKNOWLEDGEMENTS}

This work was supported by FP7 FLOWHEAD project (Fluid Optimisation
Workflows for Highly Effective Automotive Development Processes).
Grant agreement no.: 218626.

I would also like to thank professor Jacek Rokicki from the
Institute of Aeronautics and Applied Mechanics (Warsaw University of
 Technology) for encouragement and help in scientific research.

\bibliography{rei}{}
\bibliographystyle{plain}

\end{document}